\documentclass[conference]{IEEEtran}
\renewcommand\IEEEkeywordsname{Keywords}

\usepackage[utf8]{inputenc} 
\usepackage[T1]{fontenc}    
\usepackage{hyperref}       
\usepackage{url}            
\usepackage{booktabs}       
\usepackage{amsfonts}       
\usepackage{amsmath}
\usepackage{nicefrac}       
\usepackage{microtype}      

\usepackage{caption}
\usepackage{subfig}
\usepackage{graphicx}
\usepackage{multirow}

\usepackage[]{natbib}
\title{Internal node bagging}

\author{
	Shun Yi\\
	Department of Computer Science\\
	North China University of Technology\\
	\texttt{ShunYi@ncut.edu.cn} \\
}

\begin{document}
	\maketitle
	
	\begin{abstract}
		We introduce a novel view to understand how dropout works, and propose a new neural network training method named \emph{internal node bagging}, which explicitly forces a group of nodes to learn a certain feature in train time and combine those nodes to be one node in test time. It means we can use much more parameters to improve model's fitting ability in train time while keeping model small in test time. We test our method on several benchmark datasets and find it can significantly improve test performance of small models.
	\end{abstract}

	\section{Introduction}
	Neural network is a universal approximator, we can easily increase its fitting ability by adding more layers or more nodes each layer. As large labeled datasets are relatively easy to obtain now, neural network is widely used in computer vision, NLP and other domain. However, achieve state-of-the-art performance always need big models with regularization \citep{Goodfellow-et-al-2016}, or even ensemble of several models, which limits the use of neural network especially on mobile device. Although some lightweight models have been proposed recently, like \citep{Howard2017MobileNets,Zhang2017ShuffleNet}, but they rely on well-designed structures and only focus on convolutional neural network.
	
	Dropout \citep{Hinton2012Improving} is a famous regularization method in neural network training which randomly set the outputs of some hidden nodes or input nodes to zero in train time. It is commonly accepted that dropout training is similar to bagging \citep{Breiman1996Bagging}. For each training sample, dropout randomly deletes some nodes from network, and trains a thinner subnet, those subnets are trained on different samples and averaged in test time. Instead of making any real model average which will cost too much computing resource, a very simple approximate averaging method is applied by weight scaling. There are some empirical analyses show weight scaling works well in deep models \citep{Srivastava2014Dropout,Wardefarley2014An,Pham2014Dropout}. Dropout can indiscriminately and reliably yield a modest improvement in performance when applied to almost any type of model \citep{Goodfellow2013Maxout}, but may not very efficiency on small model\citep{Srivastava2014Dropout}.
	
	In this paper, we introduce a novel view to understand how dropout works as a layer-wise ensemble learning method basing on several assumptions, and propose a new training method named \emph{internal node bagging} according to our theory. We test our method on  MINIST \citep{Lecun1998Gradient}, CIFAR-10 \citep{Krizhevsky2009Learning} and SVHN \citep{Netzer2011Reading}, with fully connected network and convolutional network, find it can significantly improve test performance of small models.
	
	\section{Motivation}
	
\begin{figure}[htbp]
		\centering
		
		\subfloat[]{
			\label{Fig:zebra_net}
			\includegraphics[width=0.32\linewidth]{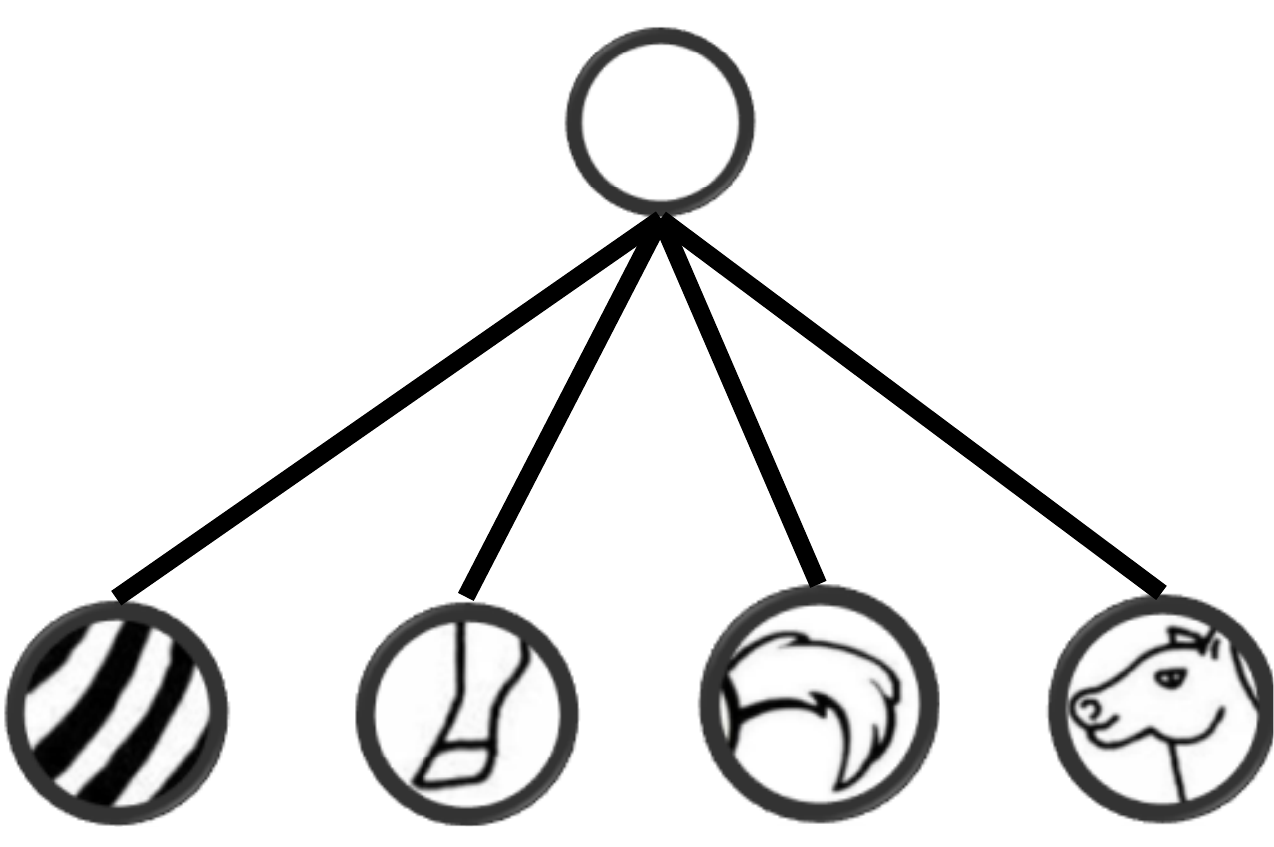}
		}
		\quad
		\subfloat[]{
			\label{Fig:grouped_zebra_net}
			\includegraphics[width=0.43\linewidth]{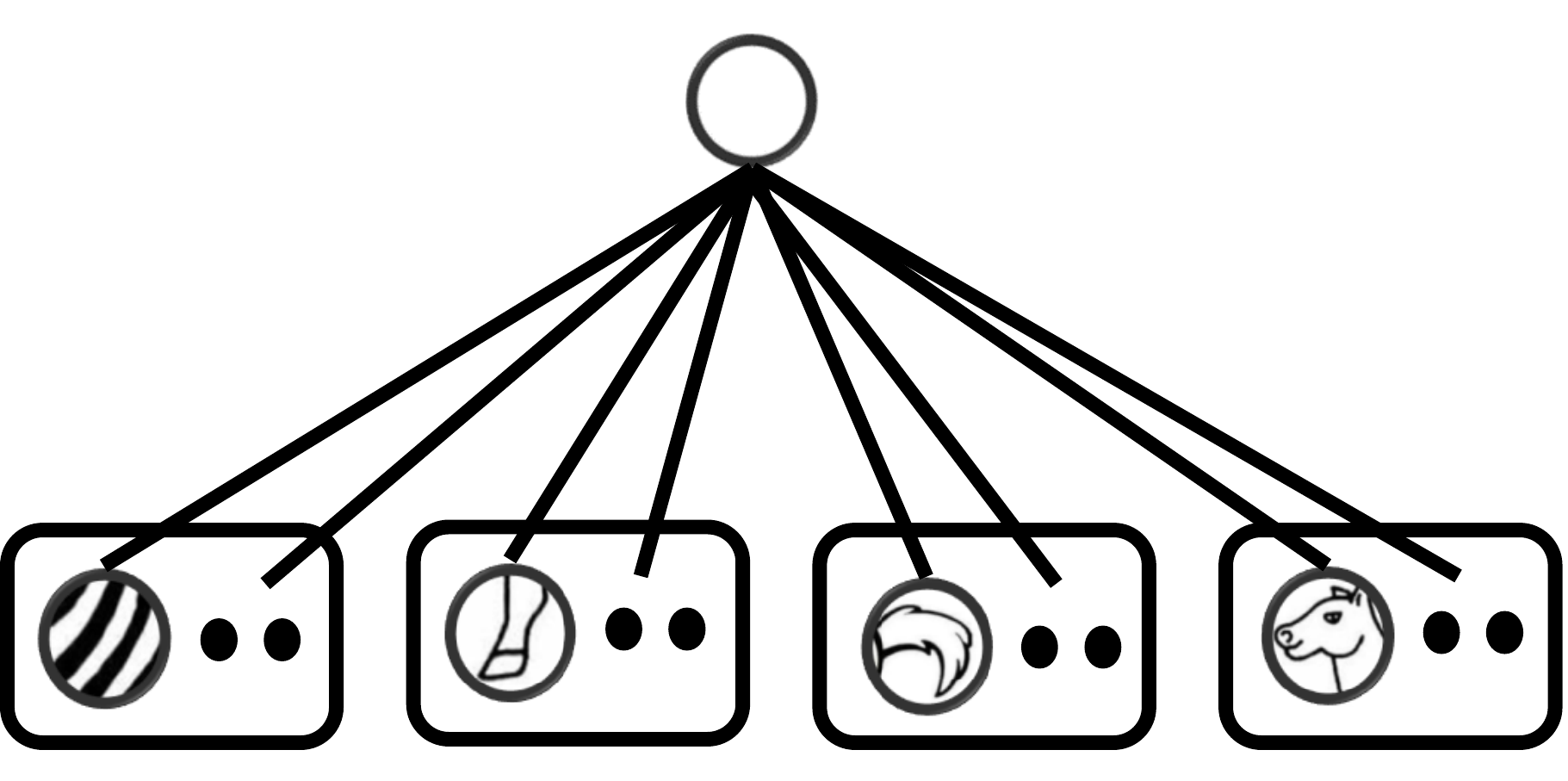}
		}	
		
		\caption{(a): An one layer fully connected network used for classifying zebra and white horse, different input nodes represent different features. (b): Similar to (a), but for every feature, there are a group of nodes represent it.}
		
	\end{figure}

	Consider a network that classify white horse and zebra in figure  \ref{Fig:zebra_net}. This simple fully connect network has 4 input nodes which represent different features belong to horse and zebra. If we apply dropout on input layer and the only difference between white horse and zebra is black-white strip, then the network absolutely cannot work if the first node which represents black-white strip is dropped.
	
	To avoid those features like black-white strip which are the key to classify different type of samples are dropped, network should learn to use more than one node to learn each of them. As we can see in figure \ref{Fig:grouped_zebra_net} which is the multi nodes per feature edition, there are 4 groups of nodes, every group contains more than one node and all of those nodes represent same feature. If drop probability is 0.5 for every node and feature's drop probability is less than 0.01, then every group should at least contain ${\log_{0.5} {0.01}}$ nodes.
	
	Although what we mentioned above is too idealized because features learned by neural network are distributed, there are still some empirical evidences support our assumption, for example, dropout training always need bigger models \citep{Srivastava2014Dropout}, shutting off a hidden neuron in dropout network can not simply remove features of input \citep{Bouthillier2015Dropout}, and there is significant redundancy in the parameterization of several deep learning models \citep{Shakibi2013Predicting}.
	
	Consider a fully connected dropout network which every feature is represented by a group of nodes like network in figure \ref{Fig:grouped_zebra_net}, the $l-1$th layer contains $k$ groups and every group has $n$ nodes, $y_{ij}^{l-1}$ is the output of the $j$th node from $i$th group of layer $l-1$, $w_{ij}^l$ and $b^l$ are the corresponding weights and biases, $m_{ij}^{l-1}$ is drop mask, $f$ is activation function, $y^l$ is the output of layer $l$. Then the forward propagation is:
	
	\begin{equation}
		y^l=f(\sum_i^k\sum_j^nw_{ij}^l(m_{ij}^{l-1}y_{ij}^{l-1})+b^l) \label{eq:prime}
	\end{equation}
	
	Because all nodes in a group represent same feature, so we can assume their weights will converge to similar value as the training process going. Let the value be $w_i^l$, so we can simplify \ref{eq:prime} to:
	
	\begin{equation}
	y^l=f(\sum_i^kw_i^l\sum_j^nm_{ij}^{l-1}y_{ij}^{l-1}+b^l) \label{eq:grouped}
	\end{equation}
	
\begin{figure}[htbp]
		\centering
		\includegraphics[width=0.5\linewidth]{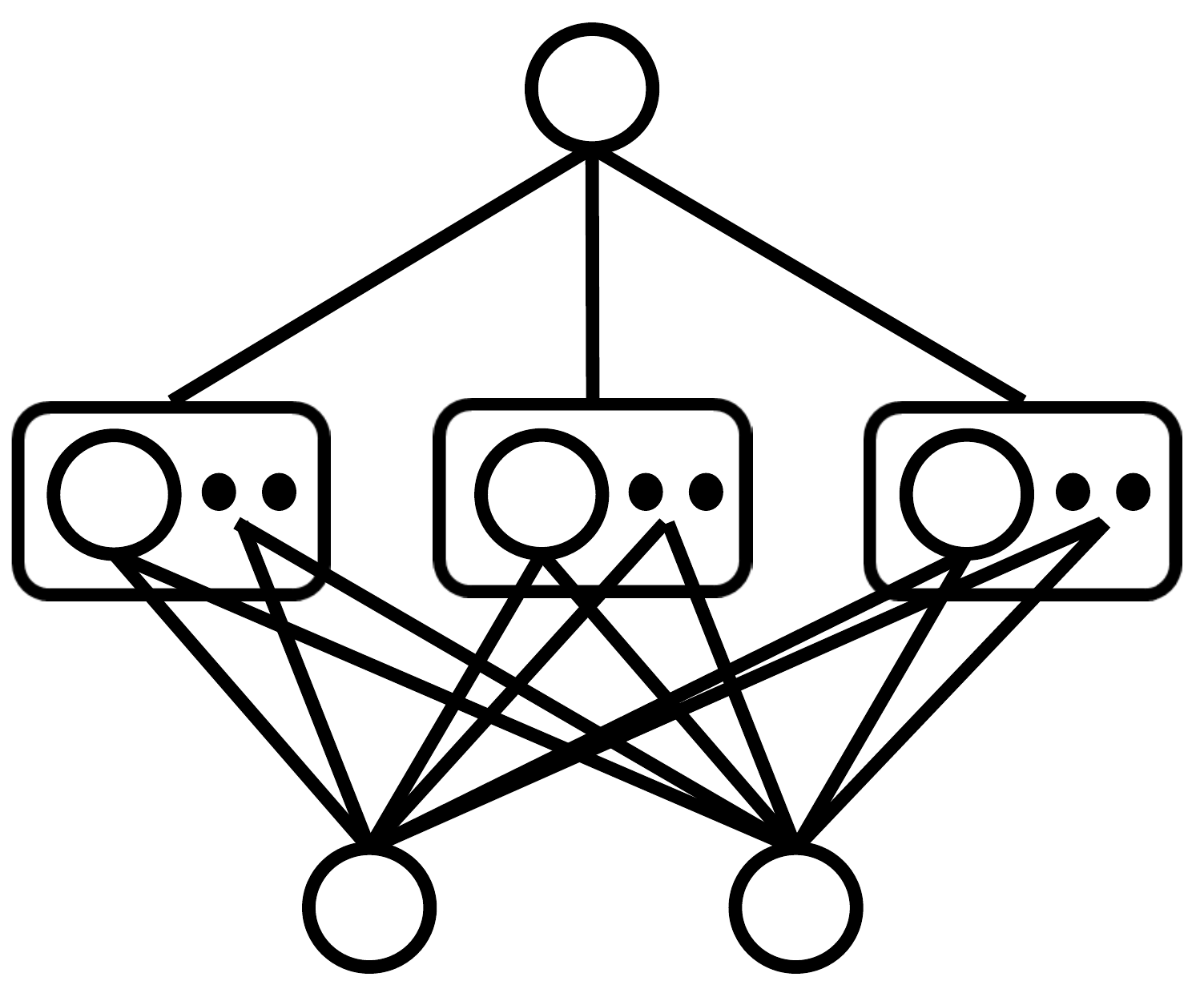}
		\caption{\emph{Internal node bagging} style network.}
		\label{Fig:internal_bagging}
	\end{figure}

	According to \ref{eq:grouped}, we can simplify neural network with dropout to network described in figure \ref{Fig:internal_bagging}, which is different to network in figure \ref{Fig:grouped_zebra_net} that every group only has one output sampled from corresponding nodes. 
	
	Network described in figure \ref{Fig:internal_bagging} is very similar to maxout \citep{Goodfellow2013Maxout}, but instead choose the biggest output in a group, it randomly samples one. Let the sampled output be $s_i^l$, it computed by:
	
	\begin{equation}
	s_i^l=\sum_j^nm_{ij}^ly_{ij}^l \label{eq:sample}
	\end{equation}
	
	Now we have a novel view to understand how dropout works as a layer-wise ensemble training method: For every feature in a layer, there are a group of nodes to learn it, next layer randomly samples a value from those nodes as the feature activation while training. In test time, weight scaling approximately let every group output the expected feature activation.
	
	We consider that if we know which nodes in a layer represent same feature, then we may be able to combine those nodes to be one node in test time which will reduce tons of parameters and computation. It can also be interpreted from opposite perspective: for every internal node in a small network, we use a group of nodes to estimate its parameters in train time. We name this method \emph{internal node bagging}.\label{Sec:understanding}
	
	There are 2 problems we should resolve, the first is how do we know which nodes represent same feature, the second is how to combine nodes to be one node. For resolving the 2 problems, we apply follow 2 tricks:
	\begin{itemize}
		\item We manually assign nodes to different groups and force them learn same feature in train time. For every group, we initialize them with same initialization and periodically compute average weights and biases and assign it to all nodes.
		\item We use relu \citep{Glorot2012Deep} in all experiments\footnote{except in the experiments of comparing the performance of different activation functions}. We assume the outputs of nodes in a group are similar as they represent same feature, so it is highly possible that those outputs distribute on the linear part of relu, and combining nodes in test time is feasible.
	\end{itemize}

	We outline detail model description in section \ref{Sec:motivation}, and experiment results in section \ref{Sec:experiments}.
	
	\section{Model description} \label{Sec:motivation}
	
	In this section, we first introduce how to combine nodes in a group to be one node and how to compute average weights, then we introduce 2 methods to sample a feature activation from a group used in our experiments.
	
	\subsection{Combine nodes} \label{Sec:comine_nodes}
	
	For given input, the expected value sampled from a group is:
	\begin{align}
	E_m(s_i^l) &= E_m(\sum_j^nm_{ij}^ly_{ij}^l)= \sum_j^nE_m(m_{ij}^l)y_{ij}^l  \label{eq:expection_1}
	\end{align}
	For all $m_{ij}^l$ in a group, they obey same distribution, so, let:
	\begin{equation}
		E_m(m_{ij}^l)=E_m(m_i^l)
	\end{equation}
	Then, we can simplify \ref{eq:expection_1} to:
	\begin{equation}
	E_m(s_i^l) = E_m(m_i^l)\sum_j^ny_{ij}^l  \label{eq:expection_2}
	\end{equation}
	Consider\footnote{$w_{ij}^l$ here is different from it in \ref{eq:prime}, they just represent corresponding weights}:
	\begin{equation}
		y_{ij}^l = relu(w_{ij}^ly^{l-1}+b_{ij}^l)
	\end{equation}
	Assume for all $y_{ij}^l$ in a group, they distribute on the linear part of relu, so:
	\begin{equation}
	E_m(s_i^l) = relu((E_m(m_i^l)\sum_j^nw_{ij}^l)y^{l-1}+E_m(m_i^l)\sum_j^nb_{ij}^l)  \label{eq:expection_4}
	\end{equation}
	We combine nodes in a group to be one node in test time according to \ref{eq:expection_4}, which is equal to weight scaling if every group only contain one node.
	
	\subsection{Compute average weights} \label{Sec:average weights}
	
	Consider a layer $l$ and group $i$, average weights are $w^{avg}$, average biases are $b^{avg}$, then:
	\begin{equation}
		nE_m(m_i^l)relu(w^{avg}y^{l-1}+b^{avg})=E_m(s_i^l)
	\end{equation}
	So:
	\begin{align}
		w^{avg}&=\frac{1}{n}\sum_j^nw_{ij}^l \\
		b^{avg}&=\frac{1}{n}\sum_j^nb_{ij}^l
	\end{align}
	We periodically compute average weights and biases in a group, and assign it to all nodes in this group to force them learn same feature.
	
	\subsection{Sample methods}
	We propose 2 methods to sample an activation from a group:
	\begin{itemize}
		\item Method A: every node in a group has same probability to be sampled independently.
		\item Method B: only one node will be sampled every group.
	\end{itemize}

	If there is only one node every group, method A is equal to dropout, network apply method B is equal to standard network.
	
	\section{Experiments} \label{Sec:experiments}
	
	We evaluate our methods on MINIST \citep{Lecun1998Gradient}, CIFAR-10 \citep{Krizhevsky2009Learning} and SVHN \citep{Netzer2011Reading}. MNIST dataset consists of 28*28 pixel gray images of handwritten digits, with 60000 samples for training and 10000 samples for testing; CIFAR-10 dataset consists of 32*32 RGB images in 10-classes with 50000 images for training and 10000 for testing; SVHN dataset consists of 32*32 RGB image dataset of digits, with 73257 images for training and 26032 images for testing. 
	
	We implement our models using tensorflow, all source code is available in \url{www.github.com/Xiong-Da/internal_node_bagging_V2}. Settings shared in all experiments are listed below:
	\begin{itemize}
		\item We apply \emph{internal node bagging} on all hidden layers.
		\item All sample probability used in method A is 0.5.
		\item We don't use any other normalization method.
		\item We default apply weight average described in \ref{Sec:average weights} every 10 epochs.
	\end{itemize}
	
	 For experiments on MNIST, we use fully connected network with 2 same width hidden layers. For experiments on CIFAR-10 and SVHN, we use CNN described in table \ref{Tab:CNN}, which is modified from the "base model C" in \citep{Springenberg2014Striving}, just remove last 1*1 convolution layer, all stride in convolution layer is 1, all padding is "SAME" except last 3*3 convolution layer.
	 
	 We train all models with Adam optimization algorithm \citep{Kingma2014Adam}. For experiments on MNIST, we train first 100 epochs with learning rate 1e-3, and train another 100 epochs with learning rate 1e-4. For Experiments on CIFAR-10 and SVHN, we train models with initial learning rate 1e-3, and decay learning rate when validate error stop decrease untill models are converged.
	 \begin{table}[htbp]
	 	\caption{The architecture of CNN used for classification experiments on CIFAR-10 and SVHN.}
	 	\label{Tab:CNN}
	 	\centering
	 	\begin{tabular}{c}
	 		\toprule
	 		32*32 RGB image \\
	 		\hline
	 		2 layer of 3*3 conv.64 \\
	 		\hline
	 		3*3 max-pooling stride 2 \\
	 		\hline
	 		2 layer of 3*3 conv.128 \\
	 		\hline
	 		3*3 max-pooling stride 2 \\
	 		\hline
	 		3*3 conv.192 \\
	 		\hline
	 		1*1 conv.192 \\
	 		\hline
	 		global averaging over 6*6 spatial dimensions \\
	 		\hline
	 		10-way softmax \\
	 		\bottomrule
	 	\end{tabular}
	 \end{table}
	 
	\subsection{Performance on models of different size} \label{Sec:performance}
	\begin{figure*}[h]
		\centering
		
		\subfloat[]{
			\includegraphics[width=0.3\linewidth]{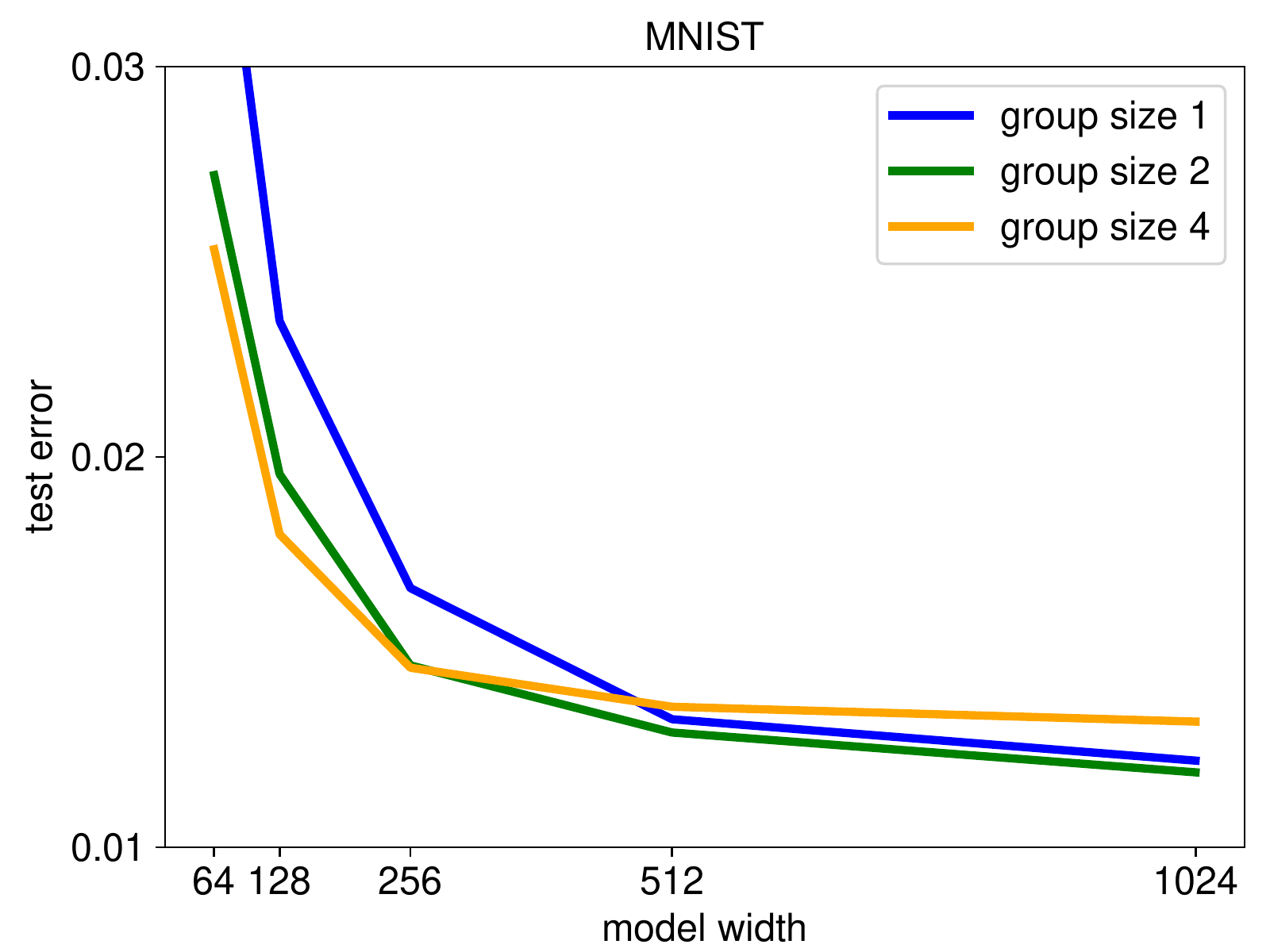}
		}
		\quad
		\subfloat[]{
			\includegraphics[width=0.3\linewidth]{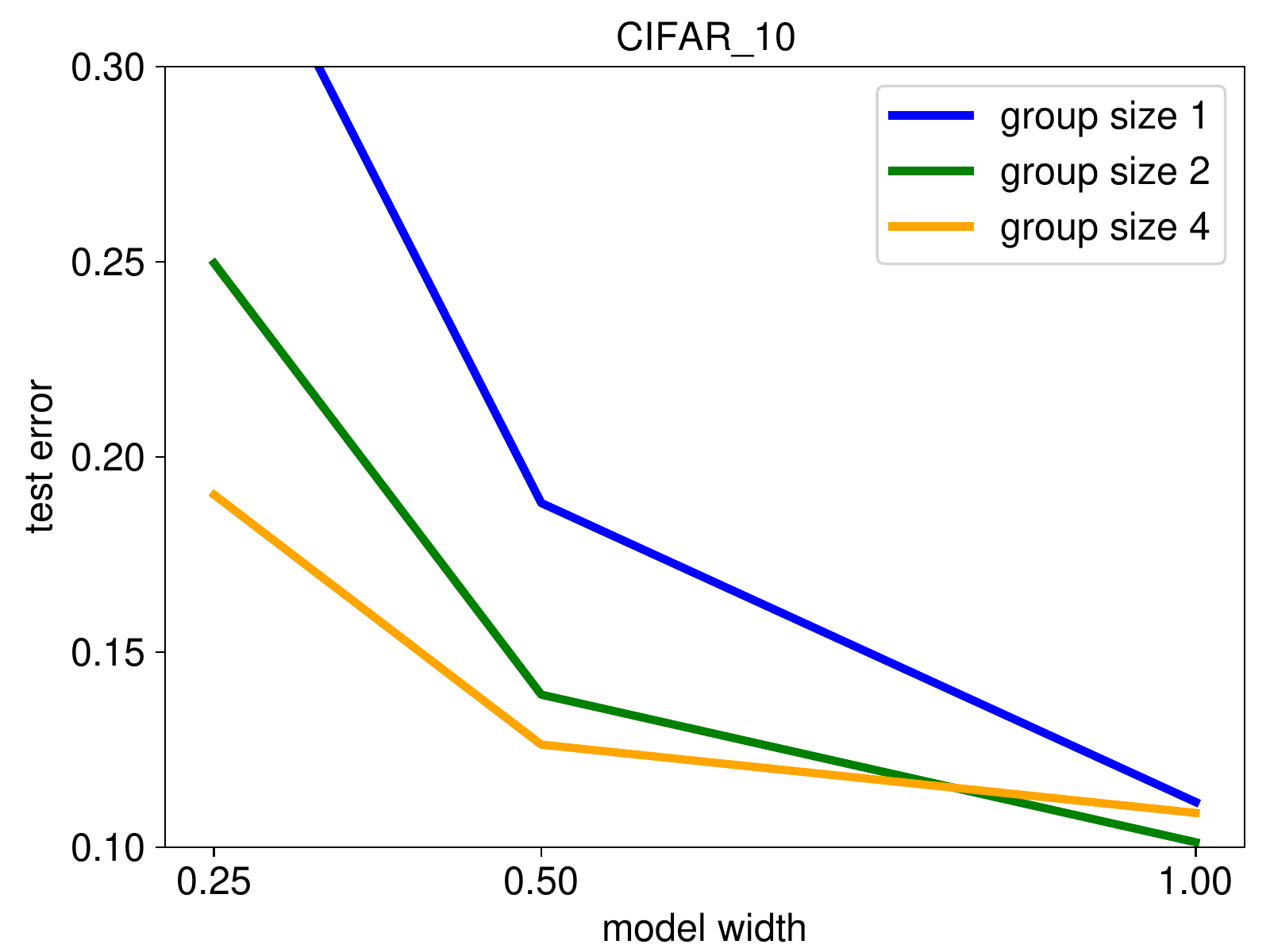}
		}
		\quad
		\subfloat[]{
			\includegraphics[width=0.3\linewidth]{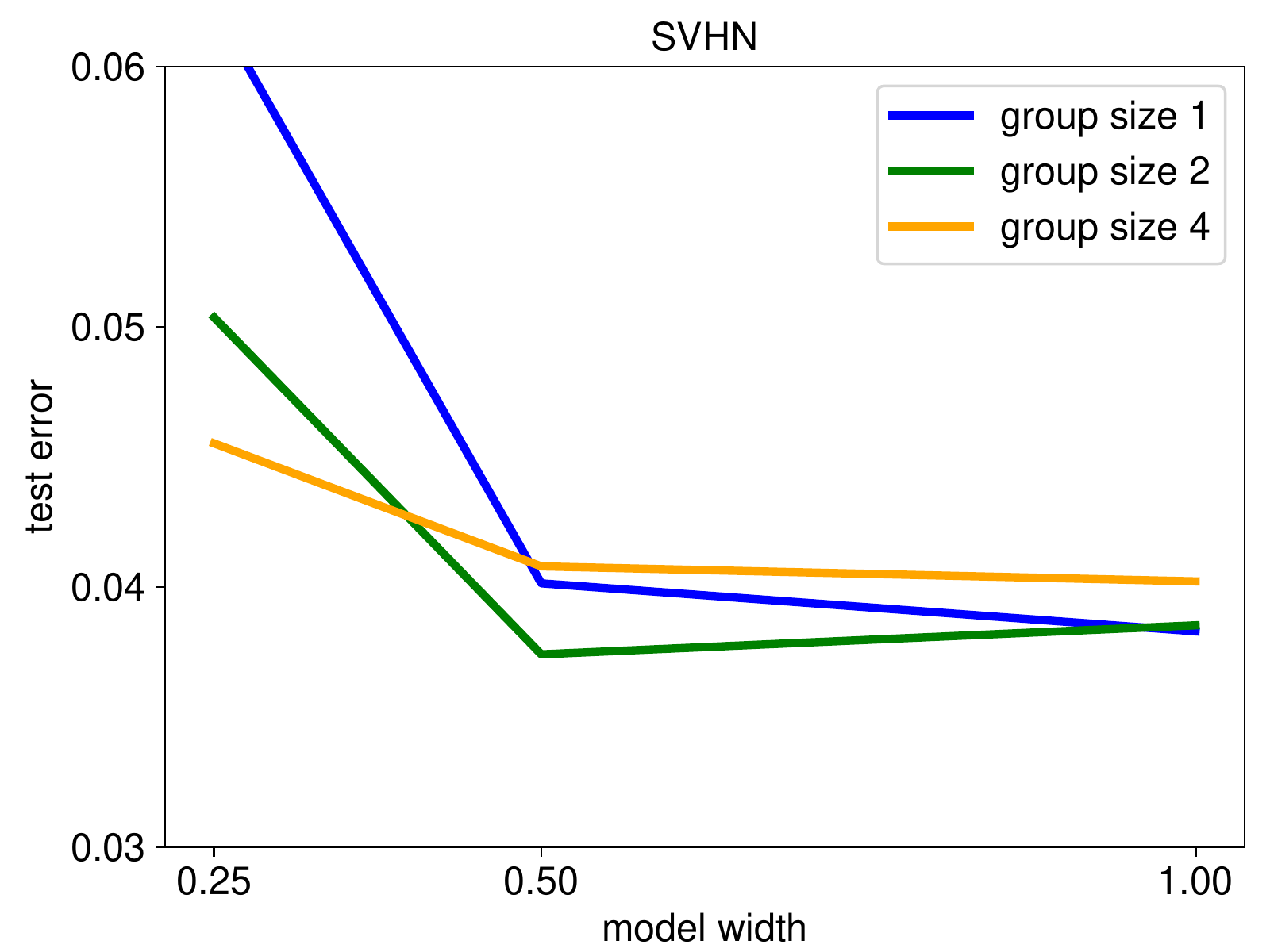}
		}
		
		\caption{Experiments on \emph{method A} with different model size. Model width in (a) means how many groups hidden layer has. Model width in (b) and (c) means the proportion of filter we use, for example, model width 0.5 means we multiply the number of filters each layer in table \ref{Tab:CNN} by 0.5.}
		\label{Fig:method_A}
	\end{figure*}
	\begin{figure*}[h]
		\centering
		
		\subfloat[]{
			\includegraphics[width=0.3\linewidth]{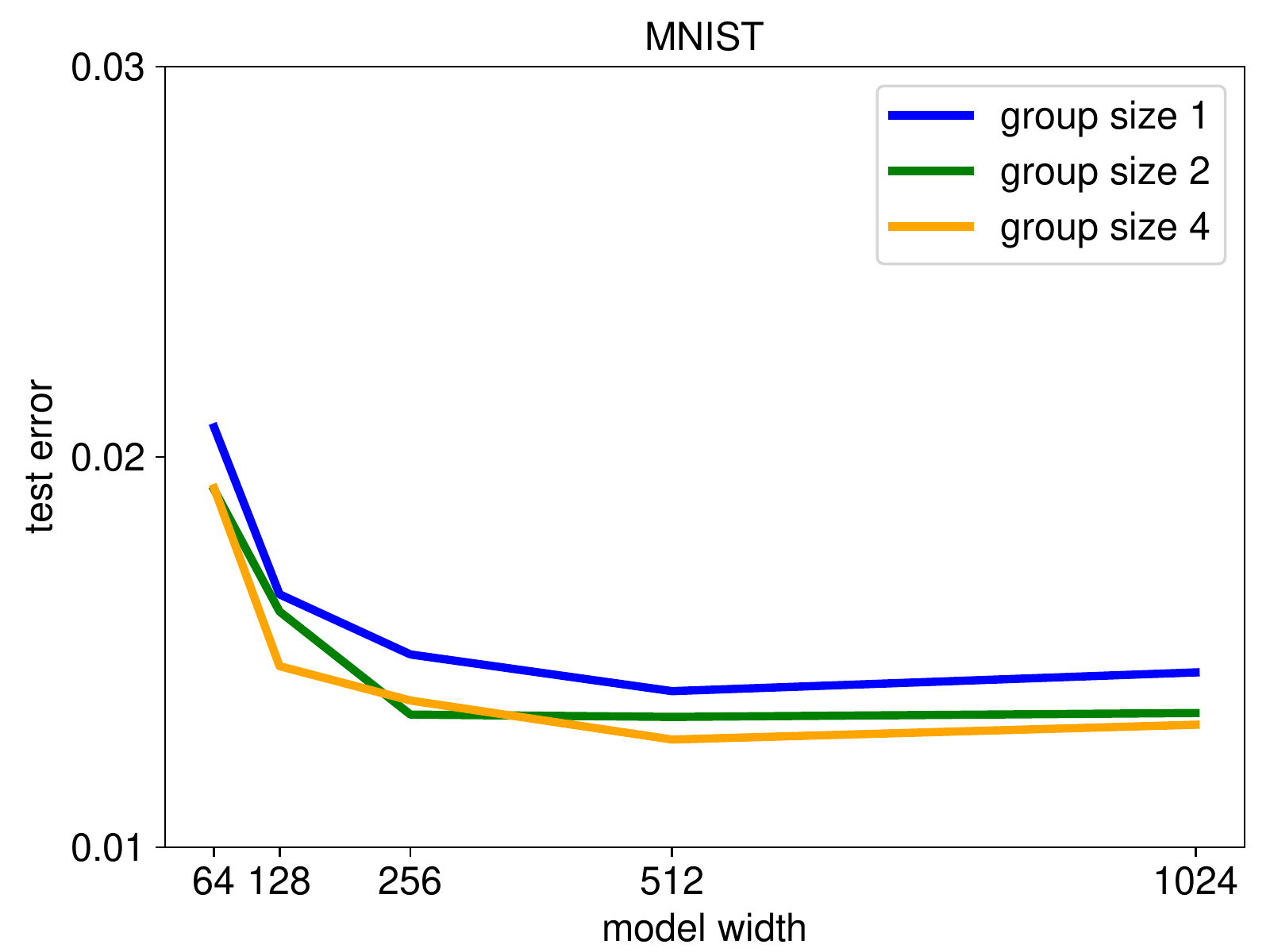}
		}
		\quad
		\subfloat[]{
			\includegraphics[width=0.3\linewidth]{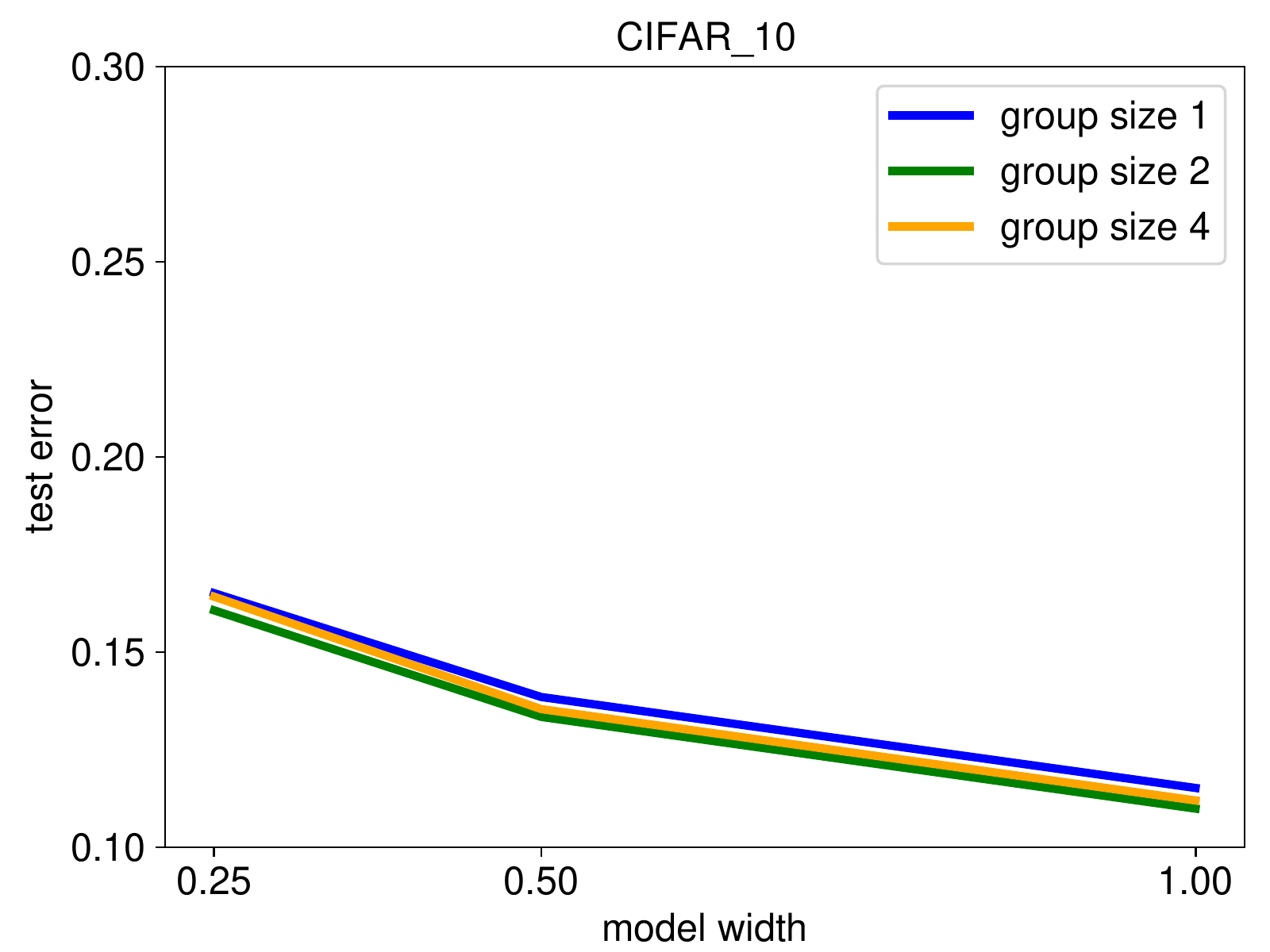}
		}
		\quad
		\subfloat[]{
			\includegraphics[width=0.3\linewidth]{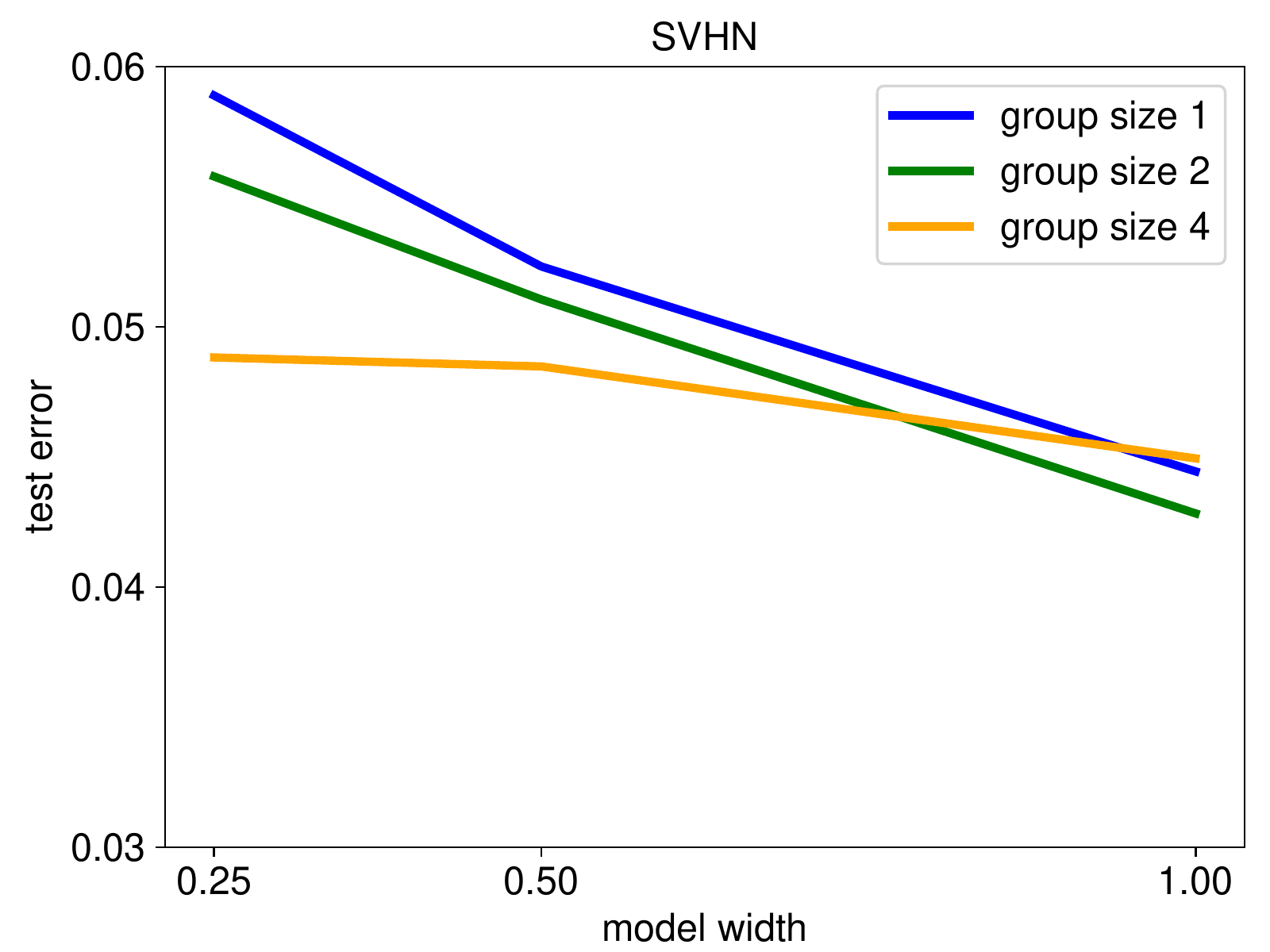}
		}
		
		\caption{Experiments on \emph{method B} with different model size.}
		\label{Fig:method_B}
		
	\end{figure*}

	In this section, we investigate the performance of our methods on models with variety size.
	
	Figures in \ref{Fig:method_A} show the performance of method A on 3 datasets. As we can see in those figures, when model size is small, increasing group size can significantly improve test performance especially on CIFAR-10 and SVHN. But as model size increase, the performance improvement start to decrease, test error of models with big group size is even worse than dropout network on SVHN (method A with group size 1 is equal to dropout).
	
	Figures in \ref{Fig:method_B} show the performance of method B. Compare to method A, method B is relatively more complex to analyze. On MNIST dataset, increasing group size can modestly improve performance both on small models and big models. On CIFAR-10 datatset, models with different group size have relatively similar performance, models with big group size perform slightly better. On SVHN dataset, method B can significantly improve performance especially on small models. 
	
	\subsection{Effect of weight average}
\begin{figure}[htbp]
		\centering
		\includegraphics[width=0.9\linewidth]{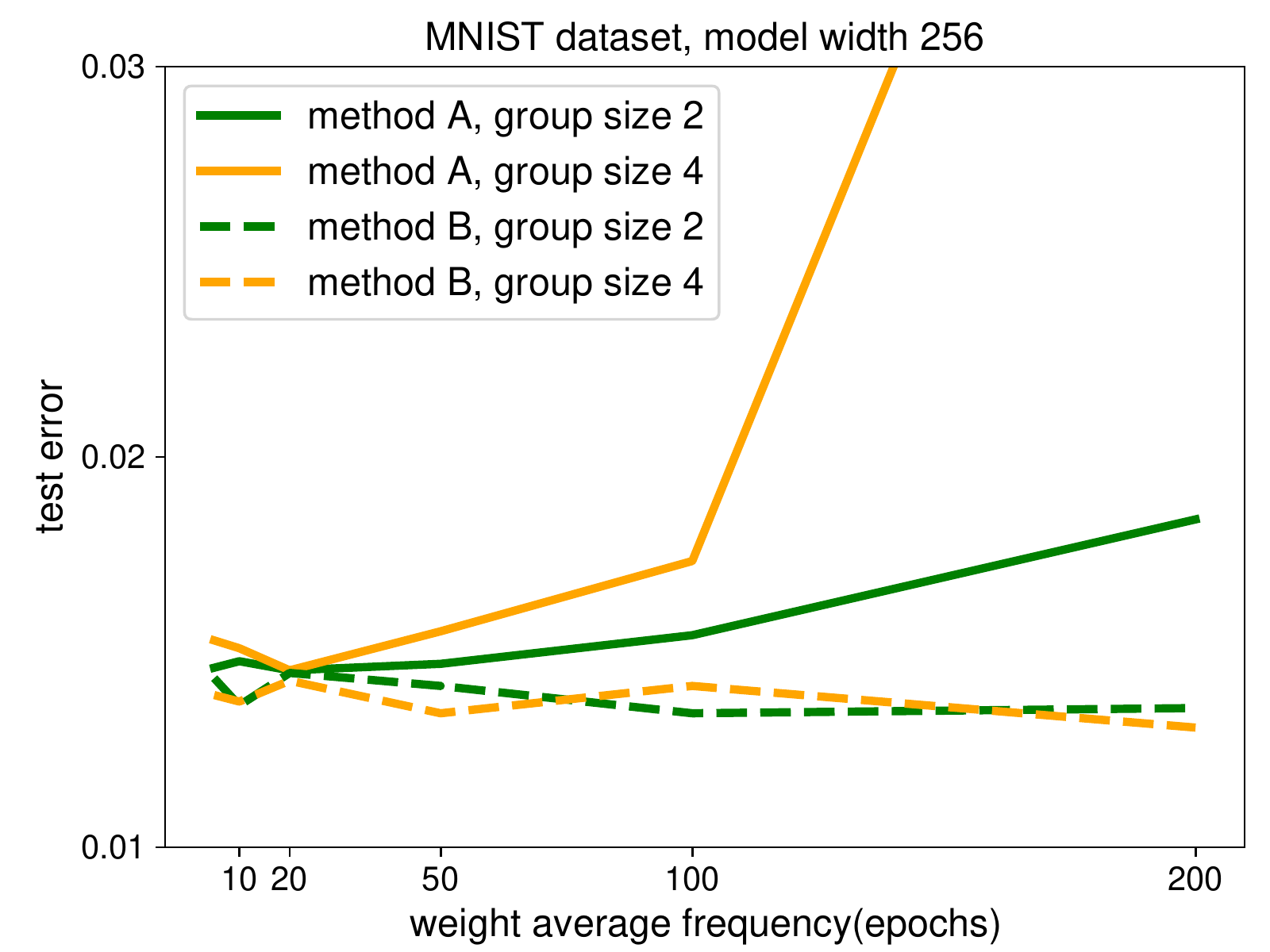}
		\caption{Analyze the effect of weight average described in \ref{Sec:average weights}.}
		\label{Fig:256_average}
	\end{figure}

	Figure in \ref{Fig:256_average} shows the effect of weight average described in \ref{Sec:average weights} on MNIST dataset with model width 256. "weight average frequency" mean train how many epochs and then apply weight average once, we only train 200 epochs on MNIST dataset, so frequency 200 means don't apply it. In our experiments, method B seemed  not sensitive to weight average frequency, but method A can't converge well without moderate frequency, especially on models with large group size.
	
	\subsection{Convergence propertie}
\begin{figure}[htbp]
		\centering
		\includegraphics[width=0.9\linewidth]{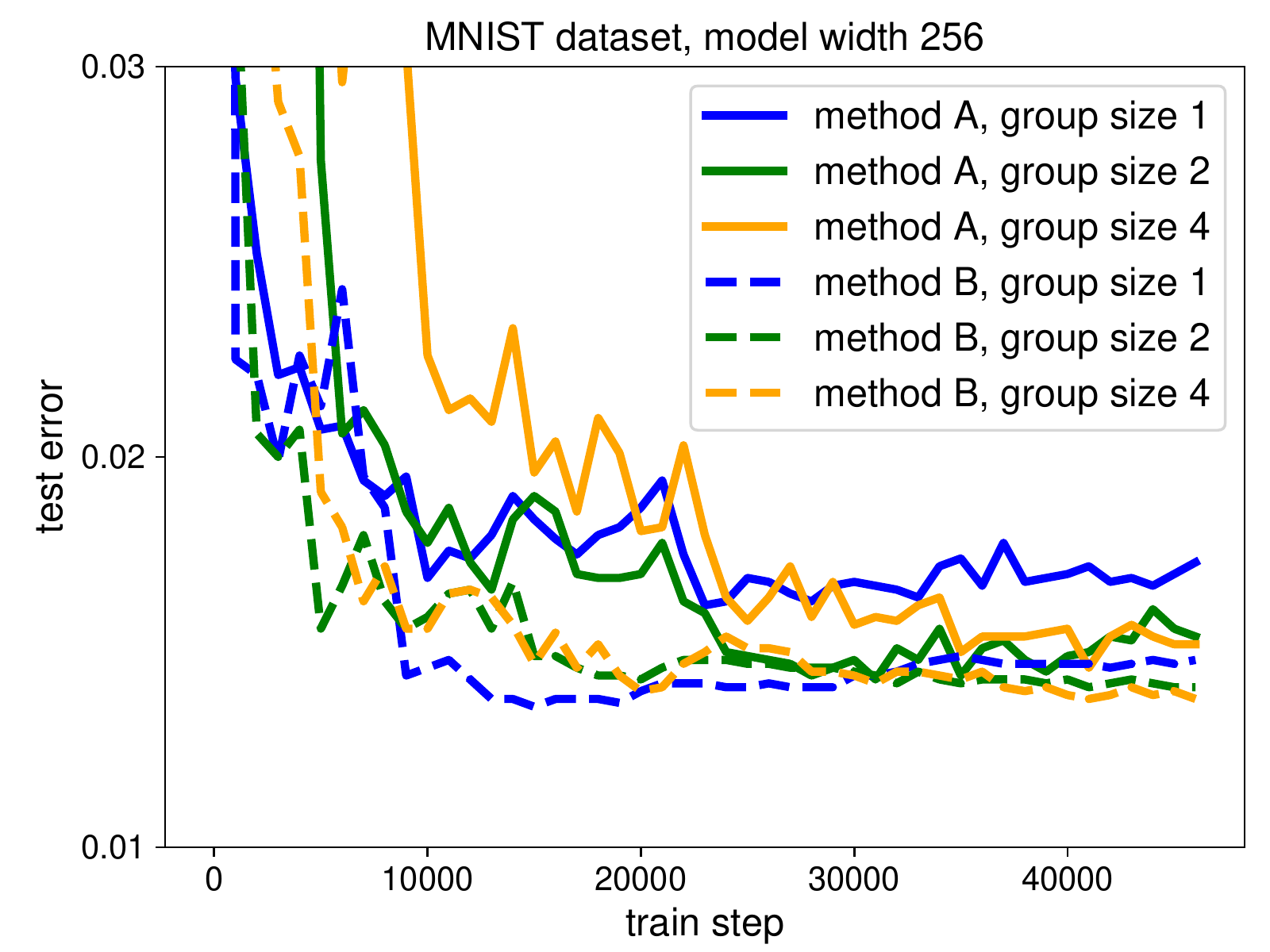}
		\caption{Analyze the convergence properties.}
		\label{Fig:256_converge}
	\end{figure}

	Figure in \ref{Fig:256_converge} shows the convergence properties of our 2 methods on MNIST dataset with model width 256. As we can see in those figure, for both methods, models with big group size do converge slower, but not slow too much.

	\subsection{Performance on different activation function}
	\begin{figure}[htbp]
		\centering
		\includegraphics[width=0.9\linewidth]{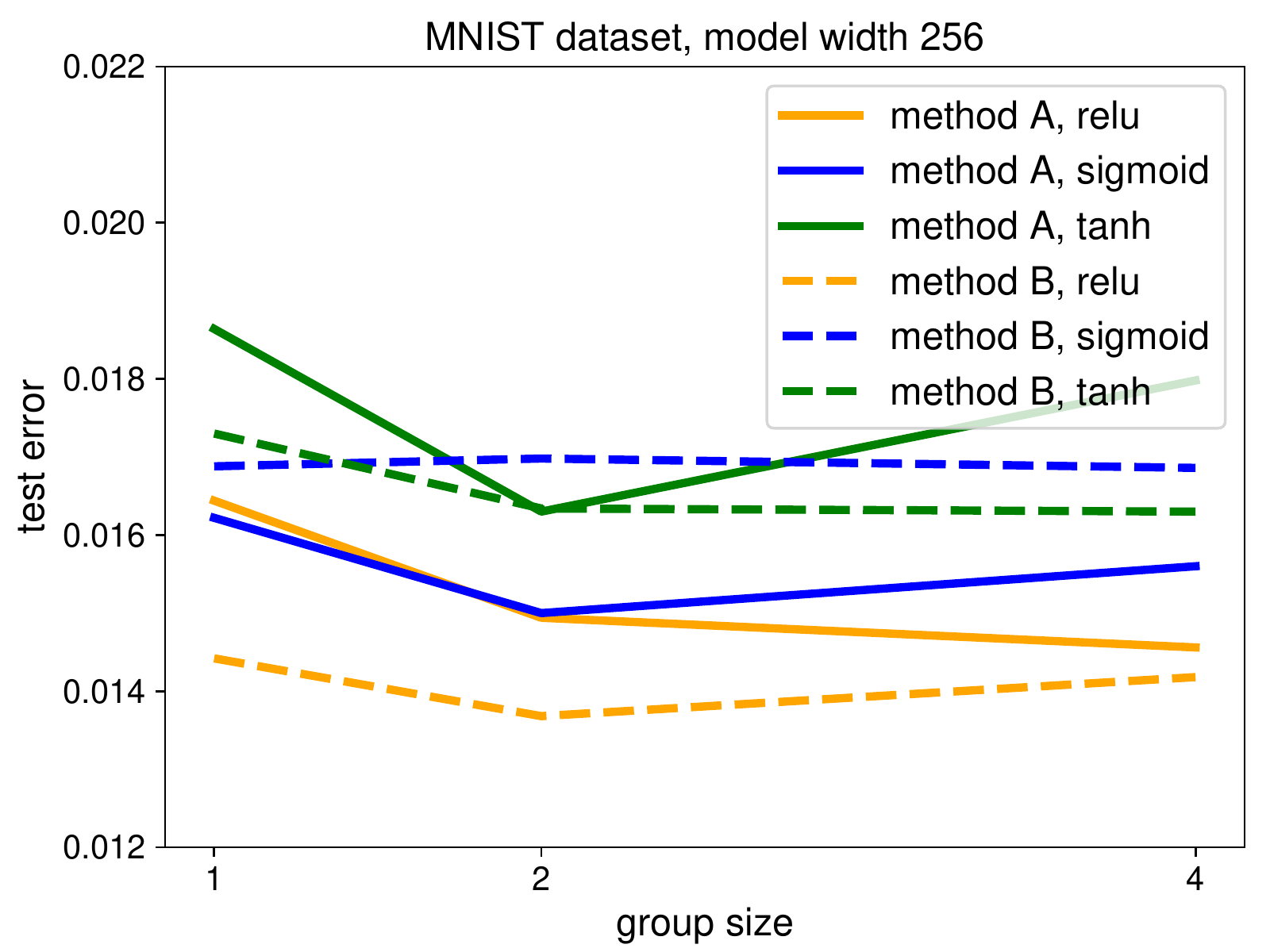}
		\caption{Analyze the performance of our methods with different activation function and group size.}
		\label{Fig:activation_functions}
	\end{figure}
	 In section \ref{Sec:comine_nodes}, we assume outputs of nodes from a group is similar and distribute on the linear part of relu as they represent same feature, and this is why combining nodes in a group to be one node in test time is feasible. In this section, we analyze that if relu is inreplaceable to our method. Figure in \ref{Fig:activation_functions} shows the experiment results of our 2 methods with 3 different activation functions on MNIST dataset. For method A, all 3 activation functions perform better when increase group size to 2, but only relu's performance keep improving when increase group size to 4. For method B, when increase group size to 2, relu and tanh perform better, but when increase group size to 4, both 2 activation function don't perform good.
	 
	 In our experiment, relu is not inreplaceable, but do perform slightly better.
	
	\section{Discussion}
	We introduced a novel view to understand how dropout works as a layer-wise ensemble learning method, and proposed a new ensemble training algorithm named \emph{internal node bagging}.
	We tested 2 sample methods in our experiments: method A can be seen as generalization of dropout, method B can be seen as generalization of standard network. For method A, increasing group size can significantly improve test performance on small models. For method B, increasing group size can moderately improve performance both on small models and big models, but it performs quite different on 3 different datasets.
	We also introduced 2 way to understand how \emph{internal node bagging} works: the first one thinks our method is equivalent to simplify big redundant models to small models without redundancy, the second one thinks our method is equivalent to estimate the parameters of every internal node by multiple nodes in train time. It seems the second one is more reasonable basing on our experiments.
	
	\bibliography{reference.bib}
\end{document}